# Ensemble feature selection with clustering for analysis of high-dimensional, correlated clinical data in the search for Alzheimer's disease biomarkers

Annette Spooner[1]*, Gelareh Mohammadi[1], Perminder S. Sachdev[2], Henry Brodaty[2], Arcot Sowmya[1]
for the Sydney Memory and Ageing Study and the Alzheimer's Disease Neuroimaging Initiative‡

[1] School of Computer Science and Engineering, UNSW Sydney, Sydney, Australia.
[2] Centre for Healthy Brain Ageing (CHeBA), School of Psychiatry, UNSW Sydney, Sydney, Australia.

*Corresponding author: Annette Spooner (a.spooner@unsw.edu.au)

## Abstract
Healthcare datasets often contain groups of highly correlated features, such as features from the same biological system. When feature selection is applied to these datasets to identify the most important features, the biases inherent in some multivariate feature selectors due to correlated features make it difficult for these methods to distinguish between the important and irrelevant features and the results of the feature selection process can be unstable.

Feature selection ensembles, which aggregate the results of multiple individual base feature selectors, have been investigated as a means of stabilising feature selection results, but do not address the problem of correlated features. We present a novel framework to create feature selection ensembles from multivariate feature selectors while taking into account the biases produced by groups of correlated features, using agglomerative hierarchical clustering in a pre-processing step.

These methods were applied to two real-world datasets from studies of Alzheimer's disease (AD), a progressive neurodegenerative disease that has no cure and is not yet fully understood. Our results show a marked improvement in the stability of features selected over the models without clustering, and the features selected by these models are in keeping with the findings in the AD literature.

## 1. Introduction

Machine learning is increasingly being applied to healthcare datasets, which often contain groups of highly correlated features, such as features from the same biological system. Feature selection is usually employed to identify the important features and aid in interpretability of the model. But the feature importance scores produced by multivariate feature selection methods can be unstable, particularly in the presence of groups of correlated features [1]. Some feature selectors choose arbitrarily amongst correlated features [2], whilst others share the feature importance scores across all correlated features in a group [1], making it difficult to separate the relevant features from those that are redundant.

Feature selection ensembles, which aggregate the results of multiple individual base feature selectors, have been investigated as a means of stabilising feature selection results [3] [4] [5] [6] [7] [8] [9] [10] [11]. Univariate filters are generally used as the base feature selectors in these methods, as they are computationally inexpensive and are not affected by correlation, but they do not account for the interactions between features. Our previous research has shown [12] that multivariate feature selectors, such as penalised regression and boosted methods, can be used in feature selection ensembles. However, these methods can be subject to bias due to groups of correlated features and this issue has not been addressed in the context of feature selection ensembles. If feature selection ensembles are to be successful in providing stable feature selection, then they must be able to handle correlated features.

The focus of this work is to build on our previous work [12] where we developed data-driven thresholds for ensemble feature selectors, by presenting a novel framework for ensemble feature selection that handles groups of correlated features. This framework clusters the features using agglomerative hierarchical clustering [13], selects the best feature from each cluster and then applies ensemble feature selection to the resulting set of weakly correlated features. Both the number and composition of the clusters is determined from the data and does not need to be defined apriori. Sparse feature selectors, which select a subset of relevant features, and those that return a score for each feature, are examined in this context, and our results show an improvement in stability over the models without clustering.

The methods developed here were applied to two real-world Alzheimer's disease (AD) datasets to identify biomarkers for AD. Despite intensive study, researchers still do not completely understand the processes that lead to the development of AD, a progressive neurodegenerative disease that has no cure. There are more than 55 million people worldwide living with dementia today, with AD thought to account for 60-70% of those cases, and numbers are forecast to triple by 2050 [14]. The processes that lead to AD begin at least 2-3 decades before overt symptoms appear [15], presenting researchers with an opportunity to identify early biomarkers that might identify patients at risk of developing AD.

AD is a slowly progressing disease, so AD datasets typically contain censored data, as participants can develop AD after the completion of the study. Therefore, multivariate feature selectors that are capable of analysing high-dimensional, heterogeneous censored data, were used to form the feature selection ensembles, showing that this method is applicable to survival analysis, not just classification and regression. The data-driven thresholds that were developed in our previous work [12] are also applied here to identify the relevant features.

The remainder of this paper is organised as follows. In Section 2 related work on instability of feature importance scores is reviewed, including the use of hierarchical clustering with feature selection and ensemble feature selection. The framework using clustering and ensemble feature selection is introduced in Section 3 and its results are presented in Section 4. In Section 5 a discussion about the method and results is provided, and Section 6 concludes the work.

## 2. Related Work

### 2.1. Bias and Instability in Feature Importance Scores

Several authors have demonstrated that feature importance scores generated by machine learning methods can be unreliable under certain circumstances, not only leading to instability but also making it difficult to distinguish between relevant and irrelevant features.

Strobl et al. [16] showed that the Gini importance measure of Brieman's random forests is not reliable when the predictors vary in their scale of measurement or number of categories, in the case of categorical features. This measure shows a bias towards features with many categories, causing features which actually have little importance to appear as highly important. This is the result of a selection bias in the individual trees and bootstrap sampling with replacement. The effect is negated if sampling without replacement is used in the construction of the individual trees.

Genuer et al. [17] examined the behaviour of random forests permutation importance and found that the variable importance scores for each feature in a group of correlated features steadily decreased as more correlated features were added to the group. This meant that relevant features that are part of a group of correlated features could achieve a feature importance score that was as low as, or lower than, some of the irrelevant features, making it very hard to distinguish between the two.

Genuer et al. also found that as the number of features $p$ increases, variable importance scores become less stable. They found that as $p$ increases, the order of magnitude of the importance scores decreases. Furthermore, as p increases, the importance scores of some noisy variables increase. Therefore, some important variables become indistinguishable from noise.

Tolosi et al. [1] also observed bias due to the presence of groups of correlated features, not only in random forests but also in the Lasso, group Lasso and fused SVM. Like Genuer et al., they found that importance scores were distributed across all correlated features in a group, due to a shared responsibility in the model, meaning that the larger the group, the smaller the importance score for each individual feature in the group. Therefore, important features can have importance scores as low as those of irrelevant features if they are part of a large enough group of correlated features.

In addition, Tolosi et al. observed that algorithms that use simple penalties, such as the Lasso, arbitrarily select one or a few features from each group of correlated features and discard the rest. As a result, the models become unstable because small changes in the training data can lead to considerable changes in the selected subset of features.

Hooker and Mentch [18] examined the biases inherent in variable importance measures for random forests, partial dependence plots, and individual conditional expectation plots, particularly where features are correlated, and sought to explain the observed behaviour. They postulated that the problem is caused by the need for these models to extrapolate in areas where there is no data. If two variables are strongly positively correlated, then there will be no observed data in the areas where one of these values is small and the other is large, and so the model must extrapolate in those areas.

## 2.2. Hierarchical Clustering for Handling Correlated Features

Clustering is the process of automatically discovering natural groupings in data. Several authors have proposed feature selection frameworks that combine clustering with feature selection algorithms to overcome the challenges of working with correlated features.

Agglomerative hierarchical clustering [13] is a method that organises objects into a tree structure, called a dendrogram, and can be constructed such that its branches represent the nested correlation structure of the data [19]. Individual clusters are identified by cutting the branches of the tree, either statically at a fixed height or dynamically. Langfelder et al. [22] developed a dynamic method of tree cutting that analyses the shape of the dendrogram branches to produce optimal clusters. This method has been shown to identify clusters that could not have been found using the static cut method.

Park et al. [20] combined hierarchical clustering with the Lasso, averaging the variables within each cluster to define supergenes, that were then used to fit a regression model. They found that the average of positively correlated genes produced a strong feature, reducing the variance of the model.

Chavent et al. [21] proposed a methodology for combining hierarchical clustering of features with feature selection, where the composition and number of the clusters is unknown a priori. The feature selection method used was VSURF (Variable Selection Using Random Forests), proposed by Genuer et al. [17]. Each group of correlated features was replaced with a synthetic numeric variable that was a linear combination of the variables in that group. The method was applied to mixed data and successfully reduced redundancy and the dimension of the problem, improving interpretability.

The methods above perform clustering first and then feature selection on a reduced set of features. Haq et al. [22] combined clustering with feature selection in a different way. Their framework first used multiple different feature selection algorithms to rank the input features, discarding those in each ranked list below a given threshold. Then the remaining features were clustered into groups, the highest ranked feature from each cluster was selected as representative of that group and the union of all the representative features formed the final feature set.

## 2.3. Feature Selection Ensembles

Stability, or reproducibility, of feature selection can be defined as the robustness of the selected features to perturbations in the data [23]. Feature selection ensembles aggregate the results of multiple base feature selectors to produce a single subset of selected features, with the aim of reducing variance to improve the stability and predictive accuracy of the results, in the same way that bagging, boosting and stacking can improve the performance and reduce the variance of supervised learning methods [24].

Saeys et al. [3] were the first to apply ensemble techniques to feature selection methods. They compared the stability and performance of individual feature selection techniques with those of homogeneous ensembles of the same techniques created from 40 bootstrap samples and a simple linear sum of the feature rankings as an aggregator. In general, the ensemble techniques provided more robust feature subsets than a single feature selector, with similar predictive accuracy.

Similar experiments have since been conducted by other researchers, varying the feature selectors, the aggregators and the number of feature subsets in the ensemble [4] [5] [6] [7] [8] [9] [10]. The aim of these experiments was to improve the stability of the features selected and the performance of the algorithms trained on those features.

Most existing studies of feature selection ensembles use only filters as base methods, as these are the most computationally efficient methods to apply to high-dimensional data. The filters chosen most frequently include chi-squared, information gain and ReliefF [4] [10] [25]. However, the original proponents of ensemble feature selection, Saeys et al. [3], applied the technique to the support vector machine (SVM) and random forests (RF). The use of multivariate feature selectors in ensemble feature selection is rare, but a notable exception [6] examines three such feature selectors– regularised regression, a tree-based gradient boosting machine and a deep neural network – successfully using these in an ensemble form. However, none of these authors have addressed the challenges of working with correlated data.

Two key elements of feature selection ensembles are the method used to aggregate the results of the individual feature selectors, and the application of a threshold to the final feature set to separate the important from the irrelevant features.

Simple mathematical combinations, such as the mean, median or sum of the feature ranks or weights, are effective aggregation techniques and are widely used [4] [3] [7] [5]. A count of the number of times each feature is selected by the base methods is also a commonly used method [26] [5]. Intuitively, if a feature is consistently given a high ranking in different data samples or by different methods, then it is likely to be important. Wald et al. [25] carried out an extensive comparison of nine different rank aggregation techniques across twenty-six bioinformatics datasets, noting that some methods were equivalent, such as the Borda Count and arithmetic mean.

Most research on ensemble feature selection applies one or more fixed thresholds to the final feature selection in order to identify the most important features. In the case of gene rank aggregation, where the number of genes can run into the thousands or even tens of thousands, a threshold of 1% of the total number of features is common [3] [4] [5]. Various other values have been suggested, including $\log_2(n)$, where n is the total number of features [9], 5% [3], 10% and 20% [10] of the total number of features.

Data-driven thresholds are rare in the literature on ensemble feature selection. Seijo-Pardo et al. [27] experimented with the use of three different data complexity measures to set automatic thresholds. However, these measures are only applicable to classification, and not to analysis of censored data.

Our previous research [12] developed and tested several data-driven thresholds, which are also applied in this work. These methods are further extended here, clustering the data and selecting the most important feature in each cluster to produce a set of weakly correlated features prior to ensemble feature selection. The number and size of the clusters is determined automatically. In this way the biases in multivariate feature selectors can be controlled, producing more reliable feature selection results.

### 2.4. Stability Measures

Somol and Novovičová [28] studied measures of feature selection stability and noted some desirable properties. The measure should be bounded by 0 and 1 where a value of 1 should imply a high level of stability, whereas a value of 0 should imply a low level of stability. The measure should also be capable of evaluating the stability of feature sets of varying sizes.

Kuncheva's *consistency index* has been used by several authors investigating the stability of feature selection subsets [4] [5] [6], but it can only be applied to subsets of identical size. Lustgarten's *adjusted stability measure* [29] handles subsets of varying sizes, but does not fulfil the other desired properties. Somol and Novovičová's [28] *relative weighted consistency* of a set of feature subsets meets the desired properties and does not overemphasise low-frequency features.

## 3.  Methods

### 3.1.  Study Cohort

Experiments in this work were conducted on data from two AD datasets to demonstrate their general applicability – the Sydney Memory and Ageing Study [30] and the Alzheimer's Disease Neuroimaging Initiative [31]. The characteristics of both studies are summarised in Table 1.

|  | MAS | ADNI-1 |
|---|---|---|
| **Study design** | Population based cohort study | Multisite longitudinal study |
| **Sample size (n)** | 873 | 819 |
| **Number of features (p)** | 140 | 216 |
| **Censoring rate** | 93% | 47% |
| **Intervals between waves** | 2 years | After 3, 6, 12, 18, 24, 36 and 48 months |
| **Age at baseline** | 70-90 years | 55 – 90 years |
| **Number of cases of AD** | 64 | 437 |

*Table 1. Study characteristics*

#### 3.1.1.  Sydney Memory and Ageing Study (MAS)

The Sydney Memory and Ageing Study (MAS) is a population-based cohort study aimed at examining the characteristics and prevalence of mild cognitive impairment and dementia. Full details of the study can be found [30]. The MAS data set contains a diverse collection of data including demographics, genetics, cognitive data, medical history, family history, medical examination, blood test results, psychological scores and functional data. Neuropsychological test scores that were used in forming a diagnosis of AD have not been used in the models developed here to predict AD.

The experiments reported in this paper used only the baseline data, collected in the first wave of MAS. Participants from a non-English-speaking background were excluded, leaving 873 participants from the original 1037. The event of interest in the survival analysis was a diagnosis of possible or probable Alzheimer's disease, over a period of 6 years, from wave 1 to wave 4 of the study. During this period 64 people developed Alzheimer's disease, indicating a censoring rate of 93%.

The Human Research Ethics Committees of the University of New South Wales and the South Eastern Sydney and Illawarra Area Health Service granted ethics approval for the MAS study and written consent was given by all participants and informants. The MAS study and this work were carried out in accordance with the MAS Governance guidelines, which are based on relevant University of New South Wales and National Health and Medical Research Council research and ethics policies.

#### 3.1.2.  Alzheimer's Disease Neuroimaging Initiative (ADNI)

The ADNI was launched in 2003 as a public-private partnership, led by Principal Investigator Michael W. Weiner, MD [31]. The primary goal of ADNI has been to test whether serial magnetic resonance imaging (MRI), positron emission tomography (PET), other biological markers, and clinical and neuropsychological assessment can be combined to measure the progression of mild cognitive impairment (MCI) and early Alzheimer's disease (AD).

ADNI participants were aged 55 - 90 years at enrolment and were recruited from 57 sites in the United States and Canada. The ADNI data set contains data from a clinical evaluation, neuropsychological tests, genetic testing, lumbar puncture, and MRI and PET scans. Subjects who participated in ADNI phase 1 were selected for this study. The event of interest in the survival analysis was a diagnosis of probable AD, over the period of the ADNI1 study. A total of 200 participants with early AD were enrolled at the start of the study and a further 237 participants developed AD during the course of the study. Data that were used in forming a diagnosis of AD, have not been used in the models developed here to predict AD.

### 3.2. Experimental Framework

To prepare the data for the feature selection ensembles, missing data were imputed using the method of multiple imputation by chained equations in the R package *mice* [32]. Imputation was performed within the cross-validation loop.

Continuous features were normalised, by subtracting the mean and dividing by the standard deviation, and multiple values for the same measurement, e.g. blood pressure, were averaged. Levels of categorical features containing only a small number of samples were combined where possible. Further details of pre-processing steps can be found elsewhere [33].

The R[34] package *mlr* (Machine Learning in R) [35] was used as a basis to carry out the experiments, while customised code was written to construct the ensembles. All of the ensembles were constructed within a 5-fold, stratified, cross-validation framework, repeated 5 times. Random probes were generated for each subsample of the data. Experiments were performed on the computational cluster Katana, supported by Research Technology Services at UNSW Sydney [36].

### 3.3. Base Feature Selectors

Six base feature selectors, capable of selecting features from high-dimensional, heterogeneous, censored data, were chosen for these experiments. Four were sparse methods, returning a subset of important features, and two were filter methods, returning a score for each feature. The four sparse methods were penalised regression for the Cox model (specifically the LASSO [37] and the ELASTIC-NET [38]), the Cox model with gradient boosting (GLMBOOST [39]) and the Cox model with likelihood-based boosting (COXBOOST [40]). The two filter methods were the maximally selected rank statistics random forests (RANGER [41]) and a univariate Cox filter (UNI). These methods represent different styles of feature selection. The Cox model, being the only univariate method chosen, is not affected by groups of correlated features, but was included for comparison with the multivariate methods.

For the LASSO, ELASTIC-NET, GLMBOOST and COXBOOST, the absolute values of the coefficients of the features were used as feature importance scores. As the data were normalised within the cross-validation loop prior to modelling, these coefficients are meaningful importance scores. For the RANGER, the method of permutation importance was used to calculate the feature importance scores. For the univariate filter, the feature importance score was the value of the C-Index returned by a Cox Proportional Hazards model applied to each feature individually.

Further information about the functioning of these methods and the R packages used to implement them can be found [33].

Each of these feature selectors was first tested in its individual form. Within a framework of 5 repeats of 5-fold cross validation, the feature selector was applied to the training data to select relevant features. Several fixed and data-driven thresholds were applied to the results of the filter methods, but as the sparse methods already select a subset of features, no further thresholding was applied to their results. A Ridge survival analysis model was trained and tested on the reduced dataset and the performance and stability of the individual models were compared to those of the ensemble models. The Ridge was chosen because of its superior performance in previous experiments [33].

### 3.4. Ensemble Construction

Homogeneous feature selection ensembles were constructed by applying the same base feature selector to 50 bootstrapped samples of the training data, producing 50 ranked subsets of features, as shown in Figure 1. Following the methods of Spooner et al [12] a variety of aggregators and both fixed and dynamic thresholds were applied to the ranked feature subsets and the resulting feature set was used as input to a Ridge survival analysis model, to assess its accuracy. The Ridge was chosen because of its superior performance in previous experiments [33].

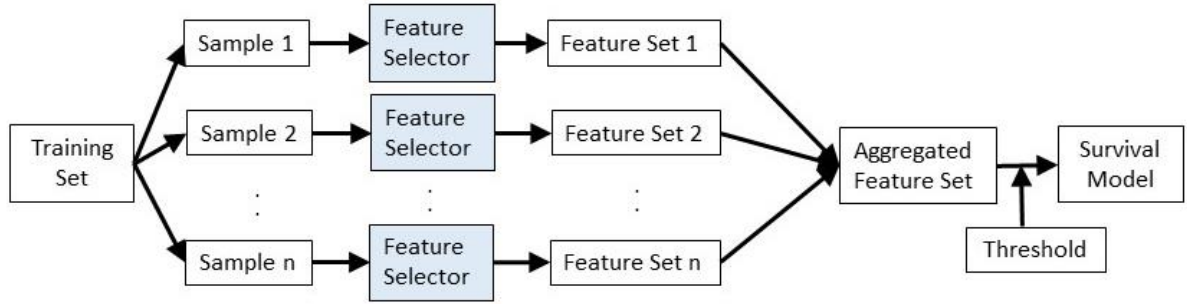

*Figure 1. A homogeneous feature selection ensemble. Sample1, Sample 2 … Sample n are randomly sampled subsets of the training data. The same feature selector is applied separately to each sample, generating n sets of selected features. An aggregator is applied to combine these feature sets into a single set and the resulting feature set is used as input to a survival model to assess its accuracy.*

### 3.5. Clustering

Agglomerative hierarchical clustering was implemented using function *hclust* in the R package *stats* v3.6.2, with the complete linkage method. Clustering was based on the Euclidean distance between spearman correlation coefficients of each pair of features.

The clustering framework is shown in part (a) of Figure 2. A dendrogram was created from the entire raw dataset, after imputing missing values. Next, clusters were identified from the dendrogram using the Dynamic Tree Cut method, in R package *dynamicTreeCut* v1.63-1. This method identifies the number and size of the clusters – they do not need to be specified apriori. A univariate Cox proportional hazards model [42] was also applied to the entire dataset, and the feature with the highest univariate Cox score in each cluster was selected to form an initial set of weakly correlated features.

The data for these features were then extracted from the original, unimputed data set, as shown in part b of Figure 2. Ensemble feature selection was applied to this data subset, using the methods described in Section 3.4.

### 3.6. Performance Metrics

The prediction accuracy of the feature selection ensembles was assessed by the value of the Concordance Index (C-Index) achieved by a Ridge penalised regression survival model trained on the features selected by the ensemble. The Ridge model was chosen for its superior performance and stability in prior experiments [33]. The C-Index measures the proportion of pairs where the observation with the higher actual survival time has the higher probability of survival as predicted by the model [43]. The performance score for the ensemble was the mean of the performance scores over five repeats of cross validation.

Each ensemble feature selector or individual method generated 25 final feature subsets during 5 repeats of 5-fold cross validation. The stability of each ensemble was measured by applying Somol and Novovičová's *relative weighted consistency* [28] to these 25 feature subsets. This metric was chosen as it is capable of evaluating feature selectors that yield subsets of varying sizes.

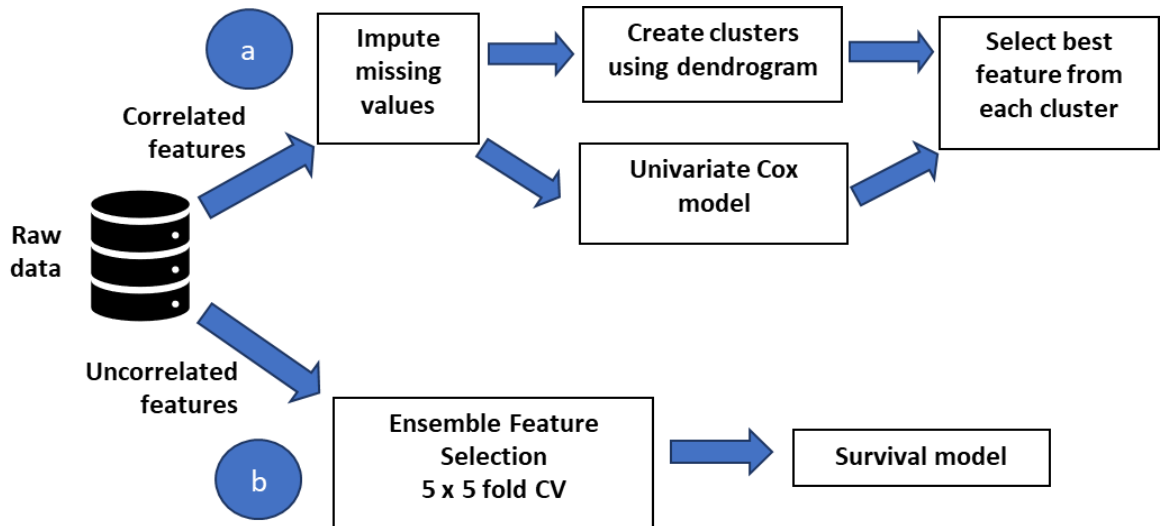

*Figure 2. Proposed framework showing selection of the best feature from each cluster (part a), followed by ensemble feature selection using uncorrelated features (part b).*

## 4. Results

The aim of this work was to build on our previous research and improve the stability of ensemble feature selectors created from multivariate base feature selectors by taking account of groups of correlated features in the data. The predictive accuracy and stability of six different feature selectors capable of handling high-dimensional, heterogeneous, right-censored data, were evaluated in individual and homogeneous ensemble form, with various aggregators and thresholds. Hierarchical agglomerative clustering was used as a pre-processing step to select a set of weakly correlated features that would minimise bias in the feature selection process. It was not necessary to define either the number or composition of the clusters apriori. The features selected by these methods were examined as possible biomarkers for AD.

The resulting models were compared using the method of Song et al. [6], where the predictive accuracy of the models was plotted against their stability. These results are shown in Figure 3 for the MAS dataset and Figure 4 for the ADNI dataset. Values further to the right of the graph indicate more stable models and values higher on the graph indicate models with a higher predictive accuracy.

A variety of thresholds was applied to the aggregated feature sets generated by the feature selection ensembles. In the individual form of the models, as the sparse feature selectors already select a subset of useful features, no further threshold was applied. In the case of the filter methods, which return a score for each feature, a threshold must always be applied, even in the individual form, to select the most useful features. Therefore, the sparse methods have a single result for the individual form of the model (represented by a pale blue star) whereas the filter methods have multiple results for the individual form of the model, one for each threshold.

Song et al. [6] determined the best model as the one with the greatest Euclidean distance from the origin in the plots of predictive accuracy vs stability. Using this method, the best ensemble model for the MAS dataset was RANGER with the RRA threshold, and the best ensemble model for the ADNI dataset was UNI with the Threshold threshold.

The main aim of using ensemble feature selection is to improve the stability of the feature selections. It is clear from the graphs in Figures 3-4 that many of the ensemble models are more stable than the individual form of the model, with a value further to the right on the graph, and all of the ensemble models based on the GLMBOOST feature selector are more stable that its individual form. In the MAS dataset, the greatest improvement in the Euclidean distance from the origin is seen in the GLMBOOST model with mean value aggregator and a threshold of 0.25. In this case the ensemble shows an increase of 0.32 or 40% over the individual form. In the ADNI dataset, the ELASTICNET model with the Threshold threshold shows the greatest improvement, with an increase of 0.16 or 14%. The graphs show that the predictive accuracy of the ensemble models is comparable to that of the individual models, therefore these increases in Euclidean distance from the origin are mainly the result of improved stability, showing that this method is beneficial.

To determine the effect of the clustering, each threshold was evaluated with and without clustering by taking the average Euclidean distance from the origin of all models that use each threshold. The results of this comparison can be seen in Table 2 and graphically in Figure 5. In almost all cases, the clustered model shows an improved performance over the equivalent model without clustering, demonstrating the benefits of this method. The exceptions are the 10% threshold on the MAS dataset, which shows a slightly worse performance and the Best Probe threshold on the MAS dataset which shows an equal performance.

The best performing threshold for the MAS dataset was the RRA threshold, which is a data-driven threshold. For the ADNI dataset, while the best performing threshold was the fixed 0.1 threshold, the 75$^{th}$ quartile, KDE ad RRA data-driven thresholds followed closely.

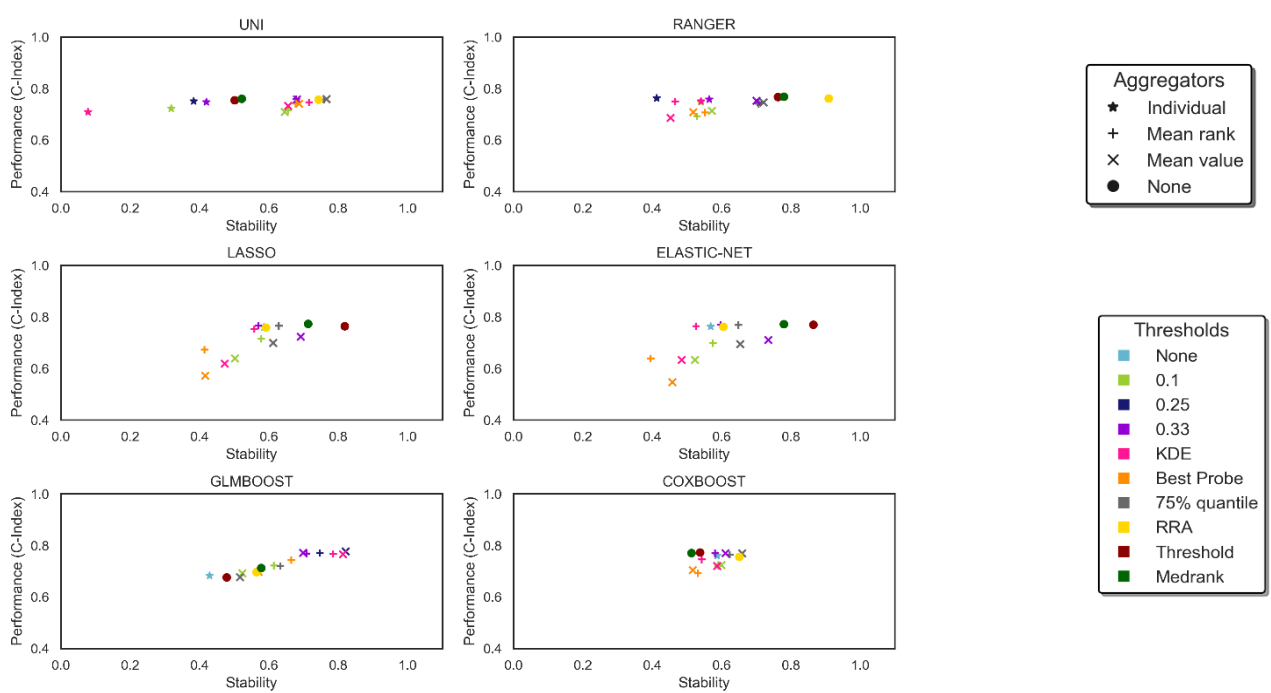

*Figure 3. Experimental results from the MAS dataset. Each plot shows performance vs stability of one feature selector. The different shapes represent different aggregators, with a star shape representing the individual form, where the model is run only once and there is no aggregation of results. The different colours represent the different thresholds applied to the models.*

|  | MAS | | ADNI | |
| --- | --- | --- | --- | --- |
|  | Non-clustering | Clustering | Non-clustering | Clustering |
| **0.1** | 0.94 | 0.90 | 1.12 | 1.22 |
| **0.25** | 0.94 | 1.00 | 1.06 | 1.14 |
| **0.33** | 0.94 | 0.99 | 1.05 | 1.13 |
| **75% quantile** | 0.96 | 0.99 | 1.07 | 1.16 |
| **Best Probe** | 0.87 | 0.87 | 1.11 | 1.14 |
| **KDE** | 0.87 | 0.92 | 1.10 | 1.15 |
| **Medrank** | 0.94 | 1.00 | 1.09 | 1.13 |
| **RRA** | 0.97 | 1.01 | 1.10 | 1.15 |
| **Threshold** | 0.97 | 1.01 | 1.08 | 1.14 |

*Table 2. Average Euclidean distance from the origin for each threshold in each dataset, with and without clustering, showing the improvement in the methods when clustering is used.*

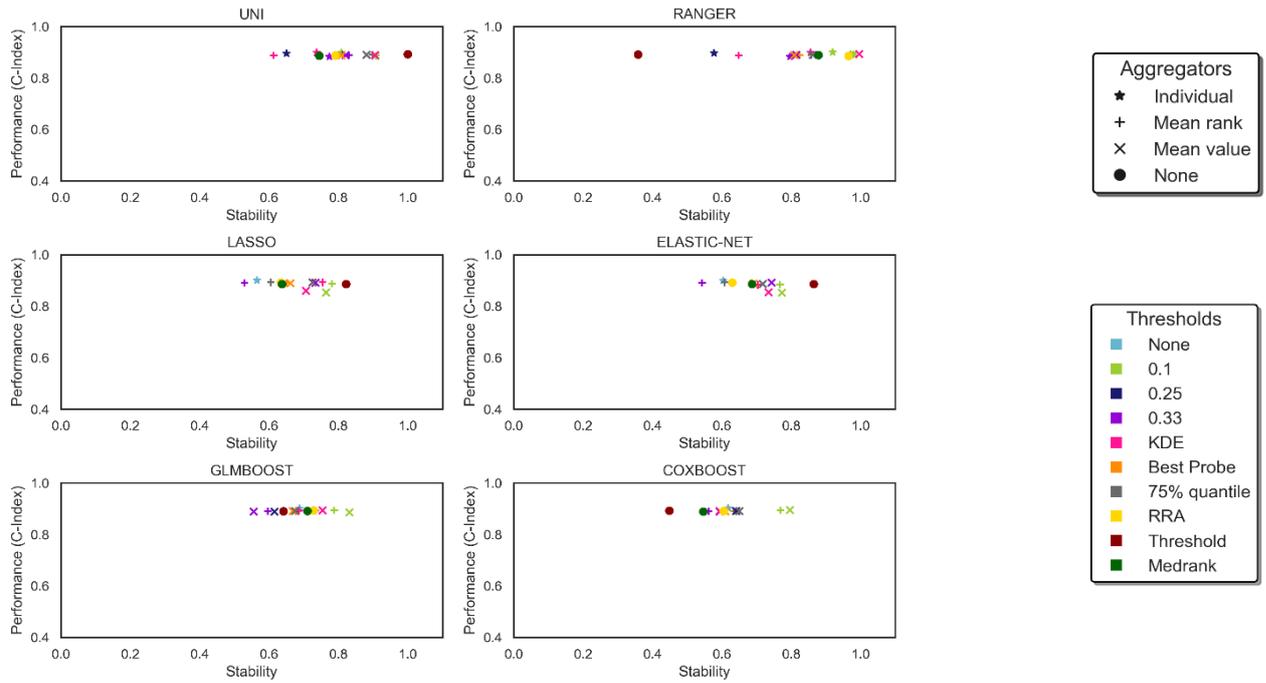

*Figure 4. Experimental results from the ADNI dataset. Each plot shows performance vs stability of one feature selector. The different shapes represent different aggregators, with a star shape representing the individual form, where the model is run only once and there is no aggregation of results. The different colours represent the different thresholds applied to the models.*

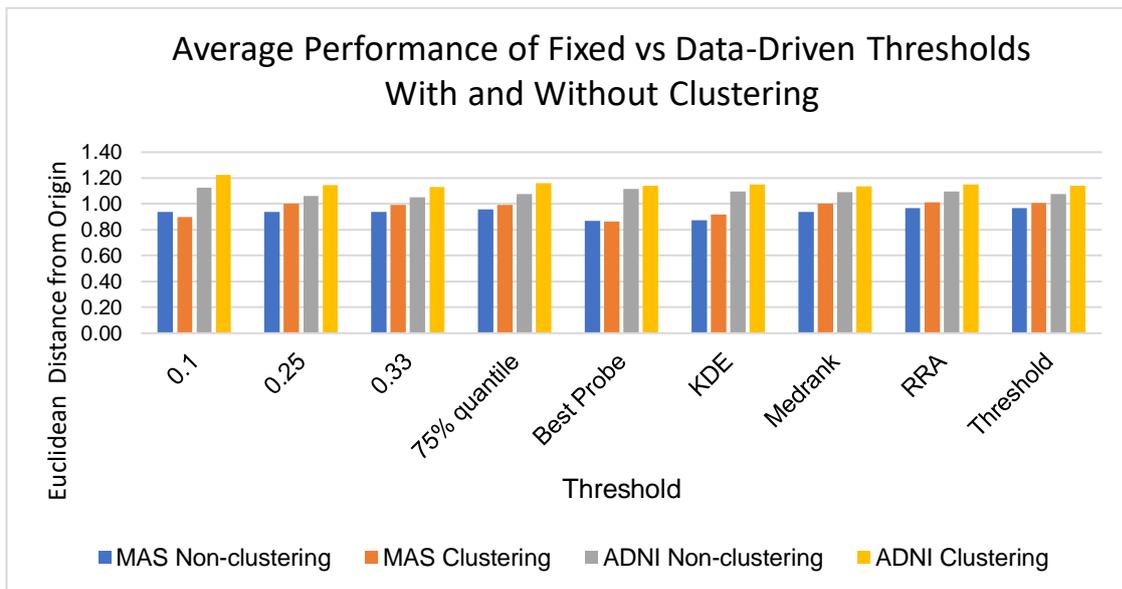

*Figure 5. Average Euclidean distance from the origin for each threshold, with and without clustering, for the ADNI and MAS datasets.*

## 4.1. Features Selected

The features selected by the ensemble feature selectors may be indicative biomarkers for AD. However, because of the stochastic nature of machine learning, even with improved stability each combination of feature selector, aggregator and threshold can still return a slightly different set of features. Therefore, a decision must be made as to which is the optimal set of selected features.

Here we decided to identify the optimal features as those selected in at least half of the top 10 performing models. Features that are selected consistently by models that perform well are likely to be important features. The features selected from the MAS dataset are described in Table 2 and those from the ADNI dataset in Table 3.

These findings are supported by the AD literature. The most important predictor of dementia is Age [44] . The risk of late-onset AD increases significantly in the presence of the E4 allele of the APOE gene [15]. Assessments using the Mini Mental State Exam [45], the General Practitioner Assessment of Cognition [46] are designed to identify dementia. Atypical changes in driving behaviours have been identified as possible early signs of mild cognitive impairment (MCI) and dementia [47] [48]. Increasing frailty, which is associated with dementia, can lead to abnormal posture [49]. Cardiovascular risk factors and obesity have been linked to an increased risk of AD [ref]. A decline in the sense of smell [50] has been suggested as an early predictor of dementia and [51]. Subjective cognitive complaints, from both participant and informant, are increasingly being recognised as predictors of progression to mild cognitive impairment and dementia [52]. The assessment of activities of daily living (ADLs) and instrumental activities of daily living (IADLs) are often used to determine the cognitive function of an individual [53].

| Feature Description - MAS |
| --- |
| Participant age at time of testing |
| Sum of scores from the Brief Smell Identification Test |
| General Practitioner Assessment of Cognition score (GPCOG) |
| Informant Questionnaire on Cognitive Decline in the Elderly (IQCODE) |
| Informant subjective cognitive complaints – total score |
| Participant subjective cognitive complaints – total score |
| Normal or abnormal posture |
| Composite variable encoding the number of major and minor at fault motor vehicle accidents in the past 18 months. |
| Framingham cardio-vascular risk score |
| Activities of daily living – total score |

*Table 2. Description of features selected by the best ensemble model trained on the MAS dataset.*

| Feature Description - ADNI |
| --- |
| Instrumental activities of daily living score |
| Mini Mental State Exam score |
| Neuropsychiatric Inventory – total score |
| Plasma neurofilament light level |
| Presence of the epsilon 4 allele of the APOE gene |

*Table 3. Meaning of features selected by the best ensemble model trained on the ADNI dataset.*

## 5. Discussion

A successful feature selection method must be both accurate and stable [6]. If accuracy alone is evaluated, the ensemble could select a different set of features at each iteration, each of which might give high predictive accuracy, but the ensemble would not be useful for knowledge discovery. On the other hand, a stable feature selection ensemble can be generated by choosing an identical set of features at each iteration, but these features may not have good predictive accuracy. Thus, the two metrics must be considered together, along with the number of features selected.

A range of different multivariate feature selectors was examined in individual and ensemble forms, with clustering used prior to ensemble feature selection to select a set of weakly correlated features. It is clear from the results presented here that the clustering step improved the stability of the models over the equivalent model without clustering. Although most previous works on ensemble feature selection have used only filters, multivariate feature selectors can also benefit by being used in an ensemble, provided the biases shown by some of these models in the presence of correlated features are controlled.

The models using clustering selected fewer features than the equivalent models without clustering, as the clustering process removes redundant features. Genuer et al [17] noted that feature selection can have two distinct goals – interpretation or knowledge discovery and prediction. In knowledge discovery, the aim is to find *all* features relevant to the target variable, including redundant features, to aid in understanding the biological processes that lead to a disease. For prediction, the aim is to find a small set of features that are sufficient to predict the target variable. The methods here are in the latter category, and one advantage of identifying a smaller yet highly predictive set of features is that they could in future lead to a reduction in the cost of testing for the disease. Current tests for AD, such as positron emission tomography (PET) scans and lumbar punctures to extract CSF, are both expensive and invasive, and an alternative set of tests with a similar predictive ability could be advantageous.

In our previous research [12] we found that the random probes threshold did not perform well with the LASSO or ELASTICNET, because a random probe was often one of the most important features selected by the LASSO, thereby eliminating other truly important features. We postulated that the problem may have been caused by correlations in the data. If that was the case, then the clustering process proposed here should have given improved results. The results did in fact improve in the ADNI dataset but not in the MAS dataset. Su et al [54] observed that false discoveries can occur very early on in the Lasso path and it seems likely that this phenomenon is occurring here. They showed that there is a trade-off between the true positive and false positive rates along the Lasso path such that it is not possible to achieve high power and a low false positive rate at the same time. This occurs even when the predictor variables are independent and there is a strong signal. They argue that this is the result of the L1 shrinkage introducing "pseudo-noise" i.e. the Lasso estimates are biased downwards when the regularization parameter is large. See Su et al [54] for more details.

As in our previous research [12], the ADNI dataset shows better performance and stability than the MAS dataset because of its lower censoring rate and controlled number of AD cases.

One disadvantage of using ensemble feature selectors is the increased computational complexity. However, with demonstrated increases in stability of up to 40%, the increased time taken to run an ensemble feature selection model may be of less importance than the greater stability that these models provide, especially for healthcare datasets where knowledge discovery is the aim. Therefore, the use of ensemble feature selection models is recommended.

Future work could apply the techniques developed here to heterogeneous ensembles of feature selectors, where a different feature selector is applied to each sample of the training data. The use of different feature selectors would introduce more diversity into the ensemble and so may produce enhanced results.

## 6. Conclusion

This work has demonstrated that ensemble feature selection combined with clustering can provide more stable, and therefore more reproducible, selections of features than individual feature selectors, even in the presence of correlated features. Results show that multivariate feature selectors of various types, both sparse methods and filters, including penalised regression, boosted methods and random forests, are suitable for use in an ensemble, provided that the biases due to the presence of correlated features are controlled. Clustering used as a pre-processing step is an effective method of choosing a set of weakly correlated features to control this bias and does not require the number or composition of the clusters to be known apriori.

The results presented here also demonstrate that ensemble feature selection is applicable not just to classification and regression but also to survival analysis, by using feature selectors that are capable of handling censored data. The use of ensemble feature selection has been expanded to a greater variety of feature selectors and a larger range of problems.

The ability to produce more stable selections of features means that clinicians can have more confidence in the results produced by machine learning models, even when the data contains highly correlated features, such as are often found in healthcare datasets. A set of features that are predictive of AD has been selected from the models developed here and these are in keeping with findings in the AD literature.

## Bibliography

[1]    L. Toloşi, T. Lengauer, Classification with correlated features: Unreliability of feature ranking and solutions,

Bioinformatics. 27 (2011) 1986–1994. https://doi.org/10.1093/bioinformatics/btr300.

[2] L. Yu, C. Ding, S. Loscalzo, Stable feature selection via dense feature groups, Proc. ACM SIGKDD Int. Conf. Knowl. Discov. Data Min. (2008) 803–811. https://doi.org/10.1145/1401890.1401986.

[3] Y. Saeys, T. Abeel, Y. Van de Peer, Robust Feature Selection Using Ensemble Feature Selection Techniques, Mach. Learn. Knowl. Discov. Databases. ECML PKDD 2008. (2008).

[4] T. Abeel, T. Helleputte, Y. Van de Peer, P. Dupont, Y. Saeys, Robust biomarker identification for cancer diagnosis with ensemble feature selection methods, Bioinformatics. 26 (2009) 392–398. https://doi.org/10.1093/bioinformatics/btp630.

[5] A. Ben Brahim, M. Limam, Ensemble feature selection for high dimensional data: a new method and a comparative study, Adv. Data Anal. Classif. 12 (2017) 1–16. https://doi.org/10.1007/s11634-017-0285-y.

[6] X. Song, L.R. Waitman, Y. Hu, A.S.L. Yu, D. Robins, M. Liu, Robust clinical marker identification for diabetic kidney disease with ensemble feature selection, J. Am. Med. Inform. Assoc. 26 (2019) 242–253. https://doi.org/10.1093/jamia/ocy165.

[7] A. Ben Brahim, M. Limam, Robust ensemble feature selection for high dimensional data sets, Proc. 2013 Int. Conf. High Perform. Comput. Simulation, HPCS 2013. (2013) 151–157. https://doi.org/10.1109/HPCSim.2013.6641406.

[8] V. Bolón-Canedo, N. Sánchez-Maroño, A. Alonso-Betanzos, Data classification using an ensemble of filters, Neurocomputing. 135 (2014) 13–20. https://doi.org/10.1016/j.neucom.2013.03.067.

[9] B. Seijo-Pardo, I. Porto-Díaz, V. Bolón-Canedo, A. Alonso-Betanzos, Ensemble feature selection: Homogeneous and heterogeneous approaches, Knowledge-Based Syst. 118 (2017) 124–139. https://doi.org/10.1016/j.knosys.2016.11.017.

[10] B. Pes, Ensemble feature selection for high-dimensional data: a stability analysis across multiple domains, Neural Comput. Appl. 3 (2019). https://doi.org/10.1007/s00521-019-04082-3.

[11] Y. Saeys, I. Inza, P. Larranaga, A review of feature selection techniques in bioinformatics, Bioinformatics. 23 (2007) 2507–2517. https://doi.org/10.1093/bioinformatics/btm344.

[12] A. Spooner, P. Mohammadi, Gelareh Sachdev, H. Brodaty, A. Sowmya, Ensemble feature selection with data-driven thresholding for Alzheimer's disease biomarker discovery, ArXiv:2207.01822. (2022).

[13] R. Xu, D. Wunsch, Survey of clustering algorithms, IEEE Trans. Neural Networks. 16 (2005) 645–678. https://doi.org/10.1109/TNN.2005.845141.

[14] World Health Organisation Fact Sheet on Dementia, (2021). https://www.who.int/news-room/fact-sheets/detail/dementia (accessed June 1, 2022).

[15] D.J. Selkoe, J. Hardy, The amyloid hypothesis of Alzheimer's disease at 25 years., EMBO Mol. Med. 8 (2016) 1–14. https://doi.org/10.15252/emmm.201606210.

[16] C. Strobl, A.L. Boulesteix, A. Zeileis, T. Hothorn, Bias in random forest variable importance measures: Illustrations, sources and a solution, BMC Bioinformatics. 8 (2007). https://doi.org/10.1186/1471-2105-8-25.

[17] R. Genuer, J.M. Poggi, C. Tuleau-Malot, Variable selection using random forests, Pattern Recognit. Lett. 31 (2010) 2225–2236. https://doi.org/10.1016/j.patrec.2010.03.014.

[18] G. Hooker, L. Mentch, Please Stop Permuting Features: An Explanation and Alternatives, ArXiv. (2019) 1–15. http://arxiv.org/abs/1905.03151.

[19] P. Langfelder, B. Zhang, S. Horvath, Defining clusters from a hierarchical cluster tree: The Dynamic Tree Cut package for R, Bioinformatics. 24 (2008) 719–720. https://doi.org/10.1093/bioinformatics/btm563.

[20] M.Y. Park, T. Hastie, R. Tibshirani, Averaged gene expressions for regression, Biostatistics. 8 (2007) 212–227. https://doi.org/10.1093/biostatistics/kxl002.

[21] M. Chavent, R. Genuer, J. Saracco, Combining clustering of variables and feature selection using random forests, Commun. Stat. Simul. Comput. 50 (2021) 426–445. https://doi.org/10.1080/03610918.2018.1563145.

[22] A.U. Haq, D. Zhang, H. Peng, S.U. Rahman, Combining Multiple Feature-Ranking Techniques and Clustering of Variables for Feature Selection, IEEE Access. 7 (2019) 151482–151492. https://doi.org/10.1109/ACCESS.2019.2947701.

[23] A. Kalousis, J. Prados, M. Hilario, Stability of feature selection algorithms: A study on high-dimensional spaces, Knowl. Inf. Syst. 12 (2007) 95–116. https://doi.org/10.1007/s10115-006-0040-8.

[24]  T.G. Dietterich, Ensemble methods in machine learning, Lect. Notes Comput. Sci. (Including Subser. Lect. Notes Artif. Intell. Lect. Notes Bioinformatics). 1857 LNCS (2000) 1–15. https://doi.org/10.1007/3-540-45014-9_1.

[25]  R. Wald, T.M. Khoshgoftaar, D. Dittman, W. Awada, A. Napolitano, An extensive comparison of feature ranking aggregation techniques in bioinformatics, Proc. 2012 IEEE 13th Int. Conf. Inf. Reuse Integr. IRI 2012. (2012) 377–384. https://doi.org/10.1109/IRI.2012.6303034.

[26]  K. Dunne, P. Cunningham, F. Azuaje, Solutions to Instability Problems with Sequential Wrapper-based Approaches to Feature Selection, Mach. Learn. (2002) 1–22. http://citeseerx.ist.psu.edu/viewdoc/download?doi=10.1.1.11.4109&rep=rep1&type=pdf.

[27]  B. Seijo-Pardo, V. Bolón-Canedo, A. Alonso-Betanzos, On developing an automatic threshold applied to feature selection ensembles, Inf. Fusion. 45 (2019) 227–245. https://doi.org/10.1016/j.inffus.2018.02.007.

[28]  P. Somol, J. Novovičová, Evaluating stability and comparing output of feature selectors that optimize feature subset cardinality, IEEE Trans. Pattern Anal. Mach. Intell. 32 (2010) 1921–1939. https://doi.org/10.1109/TPAMI.2010.34.

[29]  J.L. Lustgarten, V. Gopalakrishnan, S. Visweswaran, Measuring stability of feature selection in biomedical datasets., AMIA ... Annu. Symp. Proceedings. AMIA Symp. 2009 (2009) 406–10. http://www.ncbi.nlm.nih.gov/pubmed/20351889%0Ahttp://www.pubmedcentral.nih.gov/articlerender.fcgi?artid=PMC2815476.

[30]  P.S. Sachdev, H. Brodaty, S. Reppermund, N. a Kochan, J.N. Trollor, B. Draper, M.J. Slavin, J. Crawford, K. Kang, G.A. Broe, K. a Mather, O. Lux, The Sydney Memory and Ageing Study (MAS): methodology and baseline medical and neuropsychiatric characteristics of an elderly epidemiological non-demented cohort of Australians aged 70-90 years., Int. Psychogeriatr. 22 (2010) 1248–1264. https://doi.org/10.1017/S1041610210001067.

[31]  Mueller, Weiner, Thal, Petersen, The Alzheimer's Disease Neuroimaging Initiative, Neuroimaging Clin. N. Am. 15 (2005) 869–xii. https://doi.org/10.1016/j.pestbp.2011.02.012.Investigations.

[32]  K. van Buuren, StefGroothuis-Oudshoorn, mice : Multivariate Imputation by Chained Equations in R, J. Stat. Softw. 45 (2011). https://doi.org/10.18637/jss.v045.i03.

[33]  A. Spooner, A. Sowmya, P. Sachdev, N.A. Kochan, J. Trollor, H. Brodaty, Machine Learning Models for Predicting Dementia – a Comparison of Methods for Survival Analysis of High-Dimensional Clinical Data, Nat. Sci. Reports. (2020) 1–10. https://doi.org/10.1038/s41598-020-77220-w.

[34]  R. Team, R: A language and environment for statistical computing (Version 3.4. 2)[Computer software], Vienna, Austria R Found. Stat. Comput. (2017).

[35]  B. Bischl, M. Lang, L. Kotthoff, J. Schiffner, J. Richter, E. Studerus, G. Casalicchio, Z.M. Jones, mlr: machine learning in R, J. Mach. Learn. Res. 17 (2016) 5938–5942. https://dl.acm.org/citation.cfm?id=3053452.

[36]  Katana Computational Cluster, Https://Dx.Oi.Org/10.26190/669x-A286. (n.d.). https://doi.org/https://dx.oi.org/10.26190/669x-a286.

[37]  R.J. Tibshirani, The lasso method for variable selection in the Cox model, Stat. Med. 16 (1997) 385–395.

[38]  N. Simon, J. Friedman, T. Hastie, R. Tibrishani, Regularization Paths for Cox's Proportional Hazards Model via Coordinate Descent, (2011) 1–13. https://doi.org/10.1038/nm.2451.A.

[39]  G. Tutz, H. Binder, Boosting ridge regression, Comput. Stat. Data Anal. 51 (2007) 6044–6059. https://doi.org/10.1016/j.csda.2006.11.041.

[40]  H. Binder, M. Schumacher, Allowing for mandatory covariates in boosting estimation of sparse high-dimensional survival models, BMC Bioinformatics. 9 (2008) 1–10. https://doi.org/10.1186/1471-2105-9-14.

[41]  M.N. Wright, T. Dankowski, A. Ziegler, Unbiased split variable selection for random survival forests using maximally selected rank statistics, Stat. Med. 36 (2017) 1272–1284. https://doi.org/10.1002/sim.7212.

[42]  D.R. Cox, Regression Models and Life-Tables, J. R. Stat. Soc. 34 (1972) 187–220.

[43]  F.E. Harrell, R.M. Califf, D.B. Pryor, K.L. Lee, R.A. Rosati, Evaluating the Yield of Medical Tests, JAMA J. Am. Med. Assoc. 247 (1982) 2543–2546. https://doi.org/10.1001/jama.1982.03320430047030.

[44]  A. Mitnitski, K. Rockwood, X. Song, Nontraditional risk factors combine to predict Alzheimer disease and dementia, Neurology. 77 (2011) 227–234.

[45]  M. Folstein, S. Folstein, P. McHugh, MINI-MENTAL STATE - A practical method for grading the cognitive stats of patients for the clinician., J. Psychiatr. Res. 12 (1975) 189–198. https://doi.org/10.3744/snak.2003.40.2.021.

[46] H. Brodaty, D. Pond, N.M. Kemp, G. Luscombe, L. Harding, K. Berman, F.A. Huppert, The GPCOG: A new screening test for dementia designed for general practice, J. Am. Geriatr. Soc. 50 (2002) 530–534. https://doi.org/10.1046/j.1532-5415.2002.50122.x.

[47] S. Bayat, G.M. Babulal, S.E. Schindler, A.M. Fagan, J.C. Morris, A. Mihailidis, C.M. Roe, GPS driving: a digital biomarker for preclinical Alzheimer disease, Alzheimer's Res. Ther. 13 (2021) 1–9. https://doi.org/10.1186/s13195-021-00852-1.

[48] X. Di, R. Shi, C. Diguiseppi, D.W. Eby, L.L. Hill, T.J. Mielenz, L.J. Molnar, D. Strogatz, H.F. Andrews, T.E. Goldberg, B.H. Lang, M. Kim, G. Li, Using naturalistic driving data to predict mild cognitive impairment and dementia: Preliminary findings from the longitudinal research on aging drivers (longroad) study, Geriatr. 6 (2021) 0–9. https://doi.org/10.3390/GERIATRICS6020045.

[49] A. Mitnitski, J. Collerton, C. Martin-Ruiz, C. Jagger, T. von Zglinicki, K. Rockwood, T.B.L. Kirkwood, Age-related frailty and its association with biological markers of ageing, BMC Med. 13 (2015) 1–9. https://doi.org/10.1186/s12916-015-0400-x.

[50] S.L. Risacher, E.F. Tallman, J.D. West, K.K. Yoder, G.D. Hutchins, J.W. Fletcher, S. Gao, D.A. Kareken, M.R. Farlow, L.G. Apostolova, A.J. Saykin, Olfactory identification in subjective cognitive decline and mild cognitive impairment: Association with tau but not amyloid positron emission tomography, Alzheimer's Dement. Diagnosis, Assess. Dis. Monit. 9 (2017) 57–66. https://doi.org/10.1016/j.dadm.2017.09.001.

[51] G. Livingston, J. Huntley, A. Sommerlad, D. Ames, C. Ballard, S. Banerjee, C. Brayne, A. Burns, J. Cohen-Mansfield, C. Cooper, S.G. Costafreda, A. Dias, N. Fox, L.N. Gitlin, R. Howard, H.C. Kales, M. Kivimäki, E.B. Larson, A. Ogunniyi, V. Orgeta, K. Ritchie, K. Rockwood, E.L. Sampson, Q. Samus, L.S. Schneider, G. Selbæk, L. Teri, N. Mukadam, Dementia prevention, intervention, and care: 2020 report of the Lancet Commission, Lancet. 396 (2020) 413–446. https://doi.org/10.1016/S0140-6736(20)30367-6.

[52] M. Slavin, H. Brodaty, N. Kochan, J. Crawford, S. Reppermund, J. Trollor, B. Draper, P. Sachdev, P3-100: Predicting MCI or dementia at follow-up: Using subjective memory and non-memory complaints from both the participant and informant, Alzheimer's Dement. 7 (2011) S546–S546. https://doi.org/10.1016/j.jalz.2011.05.1540.

[53] H. Guo, A. Sapra, Instrumental Activity of Daily Living., StatPearls [Internet]. Treasure Isl. StatPearls Publ. (2021). https://www.ncbi.nlm.nih.gov/books/NBK553126/ (accessed June 28, 2022).

[54] W. Su, M. Bogdan, E. Candès, FALSE DISCOVERIES OCCUR EARLY ON THE LASSO PATH, Ann. Stat. 45 (2017) 2133–2150. https://doi.org/10.1214/16-AOS.

**Acknowledgements**

The authors acknowledge the contribution of the MAS research team and administrative assistants to this article, for their advice and collection and management of data.

Funding: The MAS study was supported by a National Health and Medical Research Council of Australia Program Grant (ID 350833). We thank the MAS study participants for their time and generosity in contributing to this research.

‡ Data from the Alzheimer's Disease Neuroimaging Initiative (ADNI) database (adni.loni.usc.edu) were used in preparation of this article. As such, the investigators within the ADNI contributed to the design and implementation of ADNI and/or provided data but did not participate in analysis or writing of this report. A complete listing of ADNI investigators can be found at: http://adni.loni.usc.edu/wp-content/uploads/how_to_apply/ADNI_Acknowledgement_List.pdf

Funding: Data collection and sharing was funded by the ADNI (National Institutes of Health Grant U01 AG024904) and DOD ADNI (Department of Defense award number W81XWH-12-2-0012). ADNI is funded by the National Institute on Aging, the National Institute of Biomedical Imaging and Bioengineering, and through generous contributions from the following: AbbVie, Alzheimer's Association; Alzheimer's Drug Discovery Foundation; Araclon Biotech; BioClinica, Inc.; Biogen; Bristol-Myers Squibb Company; CereSpir, Inc.; Cogstate; Eisai Inc.; Elan Pharmaceuticals, Inc.; Eli Lilly and Company; EuroImmun; F. Hoffmann-La Roche Ltd and its affiliated company Genentech, Inc.; Fujirebio; GE Healthcare; IXICO Ltd.; Janssen Alzheimer Immunotherapy Research & Development, LLC.; Johnson & Johnson Pharmaceutical Research & Development LLC.; Lumosity; Lundbeck; Merck & Co., Inc.; Meso Scale Diagnostics, LLC.; NeuroRx Research; Neurotrack Technologies; Novartis Pharmaceuticals Corporation; Pfizer Inc.; Piramal Imaging; Servier; Takeda Pharmaceutical Company; and Transition Therapeutics. The Canadian Institutes of Health Research is providing funds to support ADNI clinical sites in Canada. Private sector contributions are facilitated by the

**Author Contributions**

A.Sp. prepared the data, designed and ran the machine learning experiments, wrote the custom code and wrote the paper. A.So. and G.M. provided expert guidance and reviewed the manuscript. P.S. and H.B. provided data and had intellectual input into revising the manuscript.

**Competing Interests**

The authors declare that they have no competing interests.

**Corresponding Author**

Annette Spooner, UNSW Sydney, a.spooner@unsw.edu.au